\documentclass[twocolumn]{svjour3}

\usepackage{graphicx}
\usepackage{mathptmx}      
\usepackage{amssymb}
\usepackage{amsmath}
\usepackage{graphicx}
\usepackage{subfigure}
\usepackage{tabularx}
\usepackage{cite}
\usepackage{xcolor}

\journalname{Neural Computing and Applications}

\begin{document}

\title{Enhancing Trajectory Prediction using Sparse Outputs: Application to Team Sports}

\author{Brandon Victor        \and
        Aiden Nibali          \and
        Zhen He               \and
        David L. Carey
}


\institute{B. Victor, A. Nibali and Z. He \at
              Department of Computer Science and Information Technology \\
              La Trobe University, Bundoora, Australia \\
              \email{\{b.victor, a.nibali, z.he\}@latrobe.edu.au}
          \and
          D. L. Carey \at
              Discipline of Sport and Exercise Science, School of Allied Health, Human Services and Sport \\
              La Trobe University, Bundoora, Australia \\
              \email{d.carey@latrobe.edu.au}
}

\date{Received: date / Accepted: date}

\maketitle

\begin{abstract}
Sophisticated trajectory prediction models that effectively mimic team dynamics have many potential uses for sports coaches, broadcasters and spectators. However, through experiments on soccer data we found that it can be surprisingly challenging to train a deep learning model for player trajectory prediction which outperforms linear extrapolation on average distance between predicted and true future trajectories. We propose and test a novel method for improving training by predicting a sparse trajectory and interpolating using constant acceleration, which improves performance for several models. This interpolation can also be used on models that aren't trained with sparse outputs, and we find that this consistently improves performance for all tested models. Additionally, we find that the accuracy of predicted trajectories for a subset of players can be improved by conditioning on the full trajectories of the other players, and that this is further improved when combined with sparse predictions. We also propose a novel architecture using graph networks and multi-head attention (GraN-MA) which achieves better performance than other tested state-of-the-art models on our dataset and is trivially adapted for both sparse trajectories and full-trajectory conditioned trajectory prediction.
\end{abstract}

\section{Introduction}


\begin{figure}
  \centering
  \includegraphics[width=0.45\textwidth]{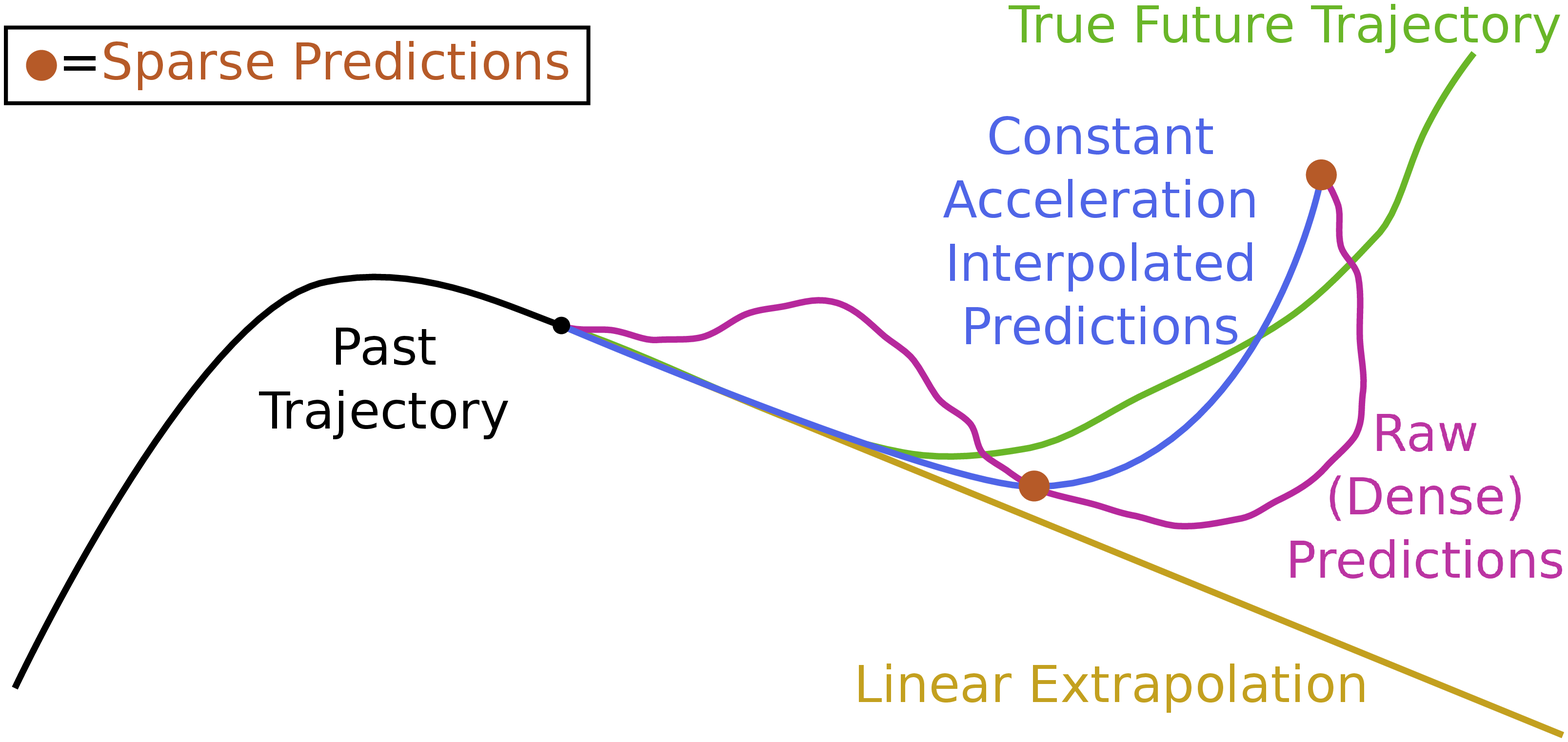}
  \caption{The intuition of using sparse predictions (brown dots) is to obtain better L2 error when the model predicts the general direction correctly, but the individual dense predictions (magenta line) do not follow physical constraints. And while linear extrapolation (yellow line) tends to match the short-term quite well, it becomes increasingly incorrect into the future. So, we utilise the sparse predictions (brown dots), along with constant acceleration interpolation (resulting in blue line), to better match the true future trajectory (green line) in both short-term and longer-term predictions.}
  \label{fig:summary}
\end{figure}

The future is uncertain. As human beings we cope with this by depending on predictions to help us: e.g. avoid walking into someone on the street, or catch a ball, or decide whether to have a picnic this Sunday. Predicting unknown positions for physical objects (e.g. people) has a wide range of applications. For example: sports play prediction \cite{Felsen2018, Yeh2019}, pedestrian tracking for autonomous cars and social robots \cite{Goldhammer2014,Ellis2009}, crowd simulation \cite{Treuille2006, Helbing1995}, and lossy compression of GPS tracking data \cite{Nibali2015, Meratnia2004}.

It is easy to hand-select weights for a linear regression model to perform linear extrapolation (i.e. to form predictions assuming constant velocity equal to the average past velocity), so we expect that even slightly more complicated multi-layer perceptron (MLP) models would therefore outperform linear extrapolation with little effort. Quite surprisingly, we found that simple baseline deep learning models such as MLPs and Recurrent Neural Networks (RNNs) were generally unable to convincingly and easily outperform linear extrapolation using conventional training on past position data. This has also been tangentially observed by others \cite{Becker2018, Felsen2018, Yeh2019}. The problem is not that these models \emph{can't} learn to outperform linear extrapolation, but rather that they are \emph{unlikely to do so} directly using standard training techniques. We do not know of any other work that seeks to address this problem in particular.

Real-world objects are subject to the laws of physics; they do not typically teleport, and their velocities change smoothly over time. Over short time intervals, the velocity does not change much, and thus the short-term motion is predicted quite well by linear extrapolation.  Instead of predicting every future step with a learned model and potentially getting these early steps or the velocity wrong, we can instead choose to make sparse predictions (e.g. one prediction per second), and then interpolate between them. By enforcing that the model can only make a few sparse predictions, we simplify the model's output space, which is easier to learn, while still being able to provide a position on every timestep. In Figure \ref{fig:summary} we summarise the intuition of our approach. We use a constant acceleration motion model for interpolation which allows gradients to be passed backwards through it. In a sense, this also embeds physical constraints about realistic movement into the loss function for these sparsely predicted positions.

In this paper, we show that training a neural network to produce sparse outputs and using constant acceleration interpolation between them results in more accurate soccer player predictions --- with respect to L2 error --- than without the interpolation. This interpolation can also be applied to any existing model by replacing the dense outputs with interpolated ones between the selected outputs. We show that this monotonically improves those models' L2 error with increasing sparsity up to 4 seconds between selected outputs.

Orthogonally to the sparse output predictions, we consider a hypothetical use case where a user (e.g. a coach) wishes to trial potential attacking manoeuvres digitally. They provide a play setup including \emph{full attacker trajectories} for the whole scenario and the past trajectories for the defenders, and the model then predicts the defenders future trajectories. This use case has also been explored in other work \cite{Felsen2018}. We verify that sparse output predictions improves this case, as well.

We also propose a novel deep learning architecture for trajectory prediction in team sports based on attention mechanisms \cite{Vaswani2017}, graph networks \cite{Battaglia2018} and other modern deep learning advances (batch normalization \cite{batchNormalisation} and skip connections \cite{ResNet}). This architecture outperforms two state-of-the-art trajectory prediction techniques -- RED \cite{Becker2018} and GVRNN \cite{Yeh2019} -- on our dataset.

Specifically, our contributions are that we:
\begin{itemize}
  \item Show that training and evaluating with sparse predictions, and using constant acceleration interpolation between, noticeably improves L2 error for most models.
  \item Show that taking an existing model's outputs, selecting a sparse subset and using constant acceleration interpolation in between monotonically improves L2 error with increasing sparsity up to 4s between selected predictions.
  \item Show that conditioning on the full trajectories of the ball/attackers to predict the future of the defenders dramatically improves L2 error, and that this is further improved by using sparse predictions.
  \item Propose and evaluate a novel architecture for trajectory prediction that uses graph networks and multi-head attention (GraN-MA) for team sports which is easily adapted for sparse predictions and full-trajectory conditioning.
\end{itemize}

\section{Related Works}

Several recent works in trajectory prediction have utilised recurrent neural network (RNN) architectures such as long short-term memory (LSTM) and gated recurrent unit (GRU) models \cite{Alahi2016, Altche2017} due to their success in language modelling \cite{Sundermeyer2012, Huang2015}, machine translation \cite{Sutskever2014, Sennrich2016} and sentiment analysis \cite{Wang2016a}. In parallel, variational auto-encoders (VAEs) \cite{Kingma2013} have been used to model the inherent uncertainty of a task, learning a distribution from which you can sample. Existing works augment RNNs with VAEs or generative adversarial networks (GANs) \cite{Goodfellow2014} to generate predictions sampled from multi-modal distributions \cite{Chung2015, Lee2017, Gupta2018}. Some recent state of the art models for language modelling and machine translation have eschewed RNNs entirely and have instead focused on attention mechanisms \cite{Vaswani2017, Radford2018} due to their simplicity and ease of training and consequently improved performance. In our work we are exploring the use of sparse trajectories, thus the main purpose is not to present the optimum architecture so we do not try a variational approach for simplicity.

Felsen et al. \cite{Felsen2018} train a conditional VAE with MLP encoders and decoders, explicitly encoding context (future) and identity (role) in addition to the past trajectories to predict the future trajectories of basketball players. As in our work, Felsen et al. noted improved predictions using a subset of players and including the full trajectories of the other players. While we encode both the future and the past trajectories at once, they explicitly encode the future positions separately. Le et. al. \cite{Le2017a} formulate the problem of trajectory prediction as imitation learning, using an LSTM for the policy and all player's positions as the environment. Their setup requires the full trajectory of attackers to predict the defenders. However, since these positions are used as the resultant environment state after performing an action, the model is only shown what the attackers did in the previous time step, rather than showing the full trajectories of the attackers at the beginning. Yeh et al. \cite{Yeh2019} combined graph networks \cite{Battaglia2018} with variational RNNs (VRNN) \cite{Chung2015} to create a generative, order-equivariant model for predicting trajectories in basketball and soccer, which they call Graph Variational RNNs (GVRNN).

In contrast to our work, the above works all produce only dense predictions. We are unaware of any prior work utilising fixed function interpolation between sparse predictions. The closest is work by Zhan et al. \cite{Zhan2018} which used an imitation learning objective to predict future trajectories with VRNNs. Similar to our work, they predicted several ``macro-intents'' as sub-goals. In contrast with our work, they then feed the macro-intents to another learned model to predict the final trajectories without any hard constraint that the macro-intents are followed. They found that they were unable to generate realistic macro-intents when training end-to-end, so these parts of their model were trained independently.

Recorded position data can be quite noisy, depending on the technologies that are used to collect it. There are a number of methods for removing this noise: such as outlier removal, a Kalman filter or CMARS \cite{Weber2012}. Our dataset has very little noise in positions, so we did not need to apply any such technique.

The above works often refer to pedestrian trajectory prediction and note the importance of interactions between people. Clearly, any method that does not include information from this complex environment cannot completely model how a person will move. The well known Social Forces model \cite{Helbing1995}, for example, assumes a target location for a pedestrian, and then models them as particles with repulsive and attractive forces to describe their path towards their target, and many other works have followed this type of energy-based modelling \cite{Luber2010, Lerner2007, Treuille2006}. These techniques have typically been focused on improving the quality of crowd simulations, in which the goal position of each simulated pedestrian is known, and each must navigate a congested area with other pedestrians. This contrasts with most modern pedestrian and sports trajectory prediction work which treats the goal location as a critical part of the prediction (as in this work).

Two common datasets used for pedestrian trajectory prediction are ETH \cite{Pellegrini2009} and UCY \cite{Lerner2007}. These two datasets are combined in the TrajNet dataset \cite{trajnet}. In these datasets the goal is to predict each pedestrian's future based on a short history of their previous 2D locations. At the time of writing, the highest scoring submission for TrajNet (with a published source) is RED \cite{Becker2018}. Compared to other models, RED is very simple: it does not model any human-human or human-environment interactions, and consists of a small GRU encoder with a single linear layer decoder. Thus, GRUs and LSTMs can outperform linear extrapolation \cite{Alahi2016, Becker2018} on specific datasets but other authors have noted difficulties using these models on different datasets. Sch{\"o}ller et al. \cite{Scholler2020} found that using constant velocity (linear extrapolation) on ETH and UCY can outperform all state-of-the-art models. Yeh et al. \cite{Yeh2019} found that RED \cite{Becker2018} performed worse than linear extrapolation on their task. Becker et al. \cite{Becker2018} and Felsen et al. \cite{Felsen2018} found that vanilla LSTMs and Social LSTMs \cite{Alahi2016} were not able to outperform linear extrapolation, either.

Initially, this seems absurd: we can easily hand-select weights for a linear regression model that will calculate the positions as for linear extrapolation, and thus we know that even a 2-layer neural network easily has enough expressive power to perform at least as well as linear extrapolation. In practice it appears that finding those weights through training is not trivial. We argue, then, that there should be deliberate effort put into developing training methods that more consistently obtain models which can match or outperform linear extrapolation. For example, consider that many trajectory prediction works predict the position offsets, rather than the absolute positions \cite{Yeh2019, Becker2018, Goldhammer2014, Hug2018, Ellis2009, Scholler2020}, but there has been little focus on the fact that this has improved the results of many different models. Goldhammer et al. \cite{Goldhammer2014} also rotated all inputs so that the input sequence always pointed in the positive x direction. Becker et al. \cite{Becker2018} noted that position offsets were simpler to model than absolute positions. Yeh et al. \cite{Yeh2019} note the similarity between predicting position offsets and skip connections \cite{He2016}. Skip connections are thought to improve training by allowing a simpler pathway through the network without losing any expressive power and can be used with most deep learning models. This is where we position sparse trajectories: a new technique that can improve trajectory prediction which is easily applied to many different models.

\section{Task Definitions}\label{sec:problem}

\begin{figure*}
\centering
\includegraphics[width=0.95\textwidth]{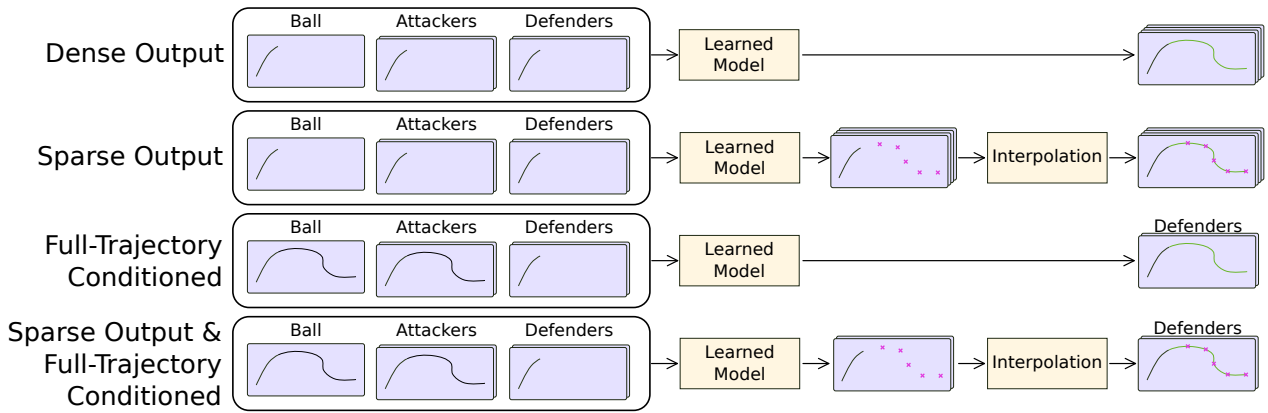}
\caption{The different inputs, outputs and layout of the different task setups. We independently choose to make the model produce sparse outputs, and/or choose to use the full trajectory of the ball and the attackers.} \label{fig:setups}
\end{figure*}

A general trajectory prediction task for team sports is a sequence to sequence problem with a single example consisting of multiple input sequences (one per player/ball), and a learned model is expected to produce a position on every future timestep for each input sequence.

\newcommand{\ourExample}{\mathbf{E}}
\newcommand{\exampleInp}{\mathbf{X}}
\newcommand{\exampleOut}{\mathbf{Y}}
\newcommand{\modelOut}{\hat{\mathbf{Y}}}

An example $\ourExample \in \mathbb{R}^{M\times T\times 2}$ in soccer consists of $M=23$ trajectories, each with $T$ timesteps of 2D coordinates: one for the ball, and 11 per team. We denote
\begin{align}
  \exampleInp = \ourExample_p &\in \mathbb{R}^{M\times n\times 2} \text{,}\\
  \exampleOut = \ourExample_f &\in \mathbb{R}^{M\times (T-n)\times 2}
\end{align}
to be the input (past) and output (future) trajectories respectively. The general trajectory prediction task is to use some model $f$ to predict $f(\exampleInp) = \modelOut$ with minimal L2 error between $\exampleOut$ and $\modelOut$.

\subsection{Sparse Trajectory Prediction}

We observed that as the window of predictions got larger, models predicting locations for each and every time step would perform relatively worse on earlier time steps (see Section \ref{sec:resultsLongTrain}). As the number of predictions increases, the models seem to trade performance on early time steps for that of later time steps because the earlier time steps don't contribute as much to the L2 error.

To address this, we reduce the number of model predictions by breaking the prediction model $f$ up into into two sub-tasks: sparse trajectory prediction $f_a$ --- in which a learned model is expected to predict only one position every n-th timestep --- and interpolation $f_b$ between those sparse predictions such that $f(\exampleInp) = f_b(f_a(\exampleInp))$. Figure \ref{fig:setups} shows the difference between this and the de facto dense approach taken by existing work. We derive a constant acceleration motion model used for interpolation $f_b$ in section \ref{sec:methodology}. We will henceforth refer to the de facto approach to trajectory prediction as dense trajectory prediction to differentiate from our proposal of sparse predictions.

\subsection{Full-trajectory Conditioned Trajectory Prediction}

\newcommand{\ball}{\mathbf{G}}
\newcommand{\attackers}{\mathbf{U}}
\newcommand{\defenders}{\mathbf{D}}

We establish a use case for trajectory prediction in which a soccer coach or performance analyst describes an attacking play on a digital chalkboard (similar to \cite{Sha2018}), and is presented with likely defences. Such a system could be used to rapidly invent and trial attack manoeuvres digitally and without risk. We separate an example $\ourExample$ into full trajectories (i.e. including all T timesteps): 1 for the ball $\ball \in \mathbb{R}^{1 \times T\times 2}$, 11 for the attackers $\attackers \in \mathbb{R}^{11 \times T \times 2}$ and 11 for the defenders $\defenders \in \mathbb{R}^{11 \times T \times 2}$. Using the subscript as before we denote $\defenders_p$ and $\defenders_f$ to be the past and future trajectories of the defenders, respectively. Then, in this case the input $\exampleInp$ consists of $\ball$, $\attackers$ and $\defenders_p$ and the output $\exampleOut$ consists of just $\defenders_f$.

This is akin to the defenders correctly guessing what the attacking team is going to do, and thus there are typically fewer plausible trajectories for them to consider in response. This is an easier task, due to the extra information, but we find the use case compelling.

The true distribution of a player's potential future trajectories is based on interactions with other players (human-human), and their position on the pitch (human-environment). But --- as observed by Becker et. al. with their RED predictor \cite{Becker2018} --- simply including that information does not necessarily improve model performance beyond what is possible without it. Then evaluating models trained for this use case can be thought of as checking that this form of additional information is effectively utilised to improve performance.

Overall, this describes two axes of variation for task setups, each with two options, resulting in 4 distinct task setups. See Figure \ref{fig:setups} for a comparison between these task setups. The first axis is sparse outputs vs dense outputs; comparing these tests if there is any improvement in performance from predicting sparse outputs and interpolating vs predicting the dense outputs directly. The second axis is standard vs full-trajectory conditioned; this will tell us if models are able to effectively utilise the extra information or not, as well as whether sparse outputs are still useful when the trajectories are easier to predict.

\section{Methodology} \label{sec:methodology}

We and others \cite{Becker2018, Scholler2020} have noted that it can be difficult to get deep learning models to outperform linear extrapolation. This is surprising because, for most neural network models, we could easily hand-select parameters such that they perform linear extrapolation exactly. The fact that some models do not find the linear extrapolation solution (or anything close), and perform significantly worse, implies that for any model/architecture there are many better global solutions that are generally inaccessible due to difficult-to-optimise loss surfaces.

Since deep learning models do not learn simple extrapolation rules, every additional output timestep in the predicted trajectory introduces a competing set of complex approximate solutions which perform well on that timestep, but not the others. When these solutions conflict each step of the optimisation is forced to trade off performance between these output timesteps (see Section 6.3). So, we hypothesise that reducing the number of output timesteps allows the model to focus capacity on fewer outputs, reducing conflicting solutions and allowing the model to perform better on those output timesteps.

By enforcing that the learned model outputs fewer future points directly, there is less that the model is required to learn. Instead of requiring the model to learn how a player moves, \emph{and} where they are going, by using constant acceleration interpolation we are only asking the model to learn where they are going. In this way we can leverage the strengths of a physical motion model with the strengths of a neural network simultaneously.

However, separating prediction into sparse prediction and constant acceleration interpolation reduces the potential expressive power of the model by reducing the resolution at which complex movements can be predicted. For example, a model that predicts only the last time step of a temporal window is only capable of describing a player trajectory where they change direction precisely once.

So there is a tradeoff here between losing expressive power and simplifying predictions to help training. Importantly, this loss of expressive power is not necessarily a detriment to the L2 error. On average, L2 error is lower if the direction is roughly correct as the specifics of precise movement are typically smaller in scale than the general direction that the player is moving. Empirically, we found that L2 error decreases if we ignore more outputs and interpolate with constant acceleration up to 4.0s between model outputs (see Section \ref{sec:resultsEvalSparsity}).

\subsection{Autoregressive models}\label{sec:methodologyAutoRegressive}

There are two general arrangements of layers in deep learning models for sequence prediction: encoder/decoder and completely autoregressive. An encoder/decoder arrangement can trivially use sparse outputs by expecting a different number of outputs in the decoder. This includes using GRUs and LSTMs since the input and output temporal resolutions are decoupled. But a completely autoregressive model only uses the output from the previous timestep --- plus any persistent state as in GRUs/LSTMs --- to calculate the output for this timestep and thus there is no real distinction within the model between the input/initialisation sequence and autoregressively generating an output. Thus it requires that the input/initialisation sequence and generation outputs have the same temporal resolution.

In this case we may choose to produce dense outputs and ignore some of them, but this does not simplify the output space; e.g. the 10th output still depends on the previous 9 outputs. Such a model may in practice perform better because the interpolation enforces smoothly changing velocity, but there appears no reason why it would train more effectively, and the added layer of indirection might actually be detrimental.

\subsection{Constant Acceleration}

Here we derive and justify our constant acceleration physical motion model for interpolation. Interpolation describes a series of points $s_t$ between the last point of the input sequence $s_0$ and the model prediction $s_T$. Motion interpolation is physically constrained: we expect a person's position to change smoothly; their initial velocity $v_0$, acceleration $a_0$, jerk (derivative of acceleration) $j_0$ etc. are described by the input sequence, and there is a predicted point $s_T$ that the player must pass through.

If we assume constant velocity $v$, then
\begin{equation}
s_t = \int v dt = vt \Rightarrow v = \frac{s_T}{T}.
\end{equation}
If we assume constant acceleration $a$, then
\begin{equation}
s_t = \iint a dt = v_0t + \frac{at^2}{2} \Rightarrow a = 2\frac{s_T - v_0T}{T^2}.
\end{equation}
If we assume constant jerk $j$, then
\begin{equation*}
s_t = \iiint j dt = v_0t + \frac{a_0t^2}{2} + \frac{jt^3}{6} \\
\end{equation*}
\begin{equation}
\;\;\;\;\;\Rightarrow j = 6\frac{s_T - v_0T - \frac{a_0T^2}{2}}{T^3}.
\end{equation}
And similarly for any $N$th derivative of position.

In reality, the movement of a human being through the physical world does not imply constant motion at any order (velocity, acceleration, etc.). However, one of the objectives of this work is to describe a simpler problem which approximates human motion over short time periods and is more amenable to training current deep learning models. A deep learning model which is allowed to predict the whole path directly can make more complex/realistic paths, but in this paper we show that they do not reliably do so, more often than not predicting wildly unrealistic paths. So, we constrain the output space with a simple physical model. We choose constant acceleration because it exhibited the best empirical results and is minimally more complex than constant velocity interpolation.

Each of these assumptions are different differentiable approximations of physical motion. Thus they can be used to train any model learned with gradient descent. A more complex motion model may be more accurate, but complexity can hinder training. By using constant acceleration interpolation, for example, and passing gradients through it, we are effectively embedding an understanding of acceleration directly into the loss. This approach of embedding prior knowledge through the structure of the problem is similar in motivation to predicting position offsets, which embeds an understanding that a future position has to be in relation to it's previous position. In both cases, this helps the model to predict realistic player movement by simplifying what the model has to predict, which makes it easier to learn.

The loss function used to train a model with sparse outputs and differentiable interpolation is the same as is used to train one without. These calculations are applied to $x$ and the $y$ independently and we do \textit{not} enforce any soft two dimensional constraints like ensuring that the 2D velocity magnitude is within a realistic distribution as this wouldn't choose a unique path and thus wouldn't be easily differentiable.

\subsection{Model Architecture}

\begin{figure}
  \centering
  \includegraphics[width=0.45\textwidth]{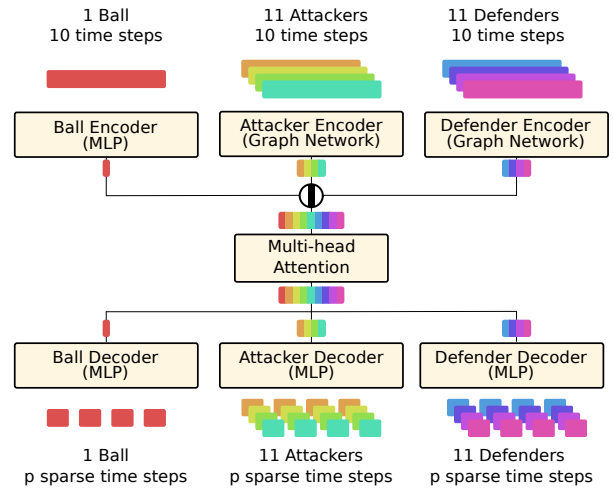}
  \caption{The GraN-MA architecture: The attacker and defender encoders are Graph Networks (GraN) with full adjacency across their team, which feed into Multihead Attention (MA) to create globally aware encodings. The predictions are separate, independent MLPs} \label{fig:architecture}
\end{figure}

We propose a novel architecture that can be applied to all of the task setups detailed in Section \ref{sec:problem}. Players on the same team are encoded together with a graph network \cite{Battaglia2018}. A globally aware encoding is then created by comparing these team aware encodings with each other and the ball encoding using multihead attention (from transformer networks \cite{Vaswani2017}). These latent features are then decoded to 2D coordinates with MLP decoders (see Figure \ref{fig:architecture}). Our Graph Network/Multihead Attention (GraN-MA) architecture is used for all task setups as adapting this architecture to sparse predictions or full-trajectory conditioning requires very few changes: adjusting the number of neurons at the output or input layers, respectively, and removing the ball and attacker decoders for full-trajectory conditioning.

Attacker and defender input trajectories are encoded using a variant of graph networks designed for modelling player interactions, as proposed by Yeh et al. \cite{Yeh2019}. This is simplified from the original graph network definition \cite{Battaglia2018} by not having any persistent state for each edge, node, or globally and always using full adjacency. More specifically,
\begin{align}
  \mathbf{e}_{i,j} &= \phi^e\left(\left[\mathbf{v}_i, \mathbf{v}_j\right]\right), \\
  \mathbf{o}_i &= \phi^v\left(\sum_j \mathbf{e}_{i,j}\right).
\end{align}
Where $\mathbf{v}_i$ are the initial node attributes, $\mathbf{e}_{i,j}$ are the edge attributes between node $i$ and $j$, and the output $\mathbf{o}_i$ are the resulting node attributes. This makes the model equivariant to the order within a team while still producing a representation per player that is aware of the other players on that team.

Standard practice for non-equivariant models is to have a canonical ordering \cite{Felsen2018, Zhan2018} for which we used team and player role metadata. Order equivariance is the property that the input order does not change the calculations, and that a change in the input order only changes the order of the output, rather than changing the output values. Models which are not order equivariant will learn biases based on the sometimes arbitrary order in which the input trajectories are presented. Both Graph Networks and Multi-head Attention are order equivariant, so they are pushed towards learning team dynamics instead of allowing them to guess based on player role. Graph networks specifically model interactions between pairs of players, and combine those pair interactions across all pairs of players. Multi-head Attention involves calculating a weighted sum over all other input embeddings; and these weights can then be used to visualise what the model is looking at for each output trajectory. Additionally, with appropriate masking, GraN-MA is theoretically robust to missing trajectories, however we did not test this.

For all MLP decoders, we use 5 hidden layers with 128 hidden units, ReLU and Batch Normalization between each of these, with skip connections across layers 2,3 and 4,5 using preactivation \cite{He2016}. Larger layers did not improve performance on our task. The MLP encoder and graph network layers $\phi^e$ and $\phi^v$ are similar to these, except they have only 3 layers in each to make them roughly the same size as the decoders. We use 4 heads for the multi-head attention \cite{Vaswani2017}; but there was little impact for different values.

\subsubsection{Baseline models}

We compare against an MLP, a simple GRU, an encoder-decoder setup using GRUs and an autoregressive CNN (using causal convolutions from \cite{Oord2016}). These baselines are not order-equivariant like GraN-MA, thus we used a canonical ordering.

The MLP baseline has the same structure as the MLP decoders, but with 2048 hidden units, chosen as the highest power of 2 that fit comfortably in RAM. This MLP baseline is much larger and slower than GraN-MA and takes a flattened vector of all players and positions and predicts a flat vector which we reshape into predictions for each time step for each player.

The simple GRU is a 2-layer GRU with 128 features (comparable to GraN-MA embedding size), the hidden state initialised with 0s, and a decoding linear layer applied on each time step to produce 2D coordinates. At each timestep, the GRU receives the positions of all players and predicts for all players simultaneously.

The encoder-decoder setup using GRUs has two distinct two-layer GRUs, each with 128 features. The encoder GRU uses the input sequence to produce a hidden state, which the decoder GRU uses to produce 2D player positions autoregressively via a linear layer applied on each time step.

Unlike \cite{Yeh2019, Becker2018}, we did not predict offsets from the last input position for any of our models, as it did not improve performance on our task.

\subsection{Loss}

The loss is averaged over trajectories and timesteps
\begin{equation}
  \mathcal{L} = \frac{1}{M\left(T-n\right)} \sum^M_m\sum^{T}_{t=n} h\left(\exampleOut^m_t, \modelOut^m_t\right).
\end{equation}
During training we use the average Huber loss \cite{Huber1964} across the 2D co-ordinates
\begin{equation}
h(\mathbf{y}, \hat{\mathbf{y}}) =
  \frac{1}{2}
  \sum_i^2
  \begin{cases}
  0.5 (y_i - \hat{y}_i)^2, & \text{if } |y_i - \hat{y}_i| < 1 \\
  |y_i - \hat{y}_i| - 0.5, & \text{otherwise.}
  \end{cases}
\end{equation}
During testing we use L2 error
\begin{equation}
h(\mathbf{y}, \hat{\mathbf{y}}) = ||\mathbf{y} - \hat{\mathbf{y}}||_2.
\end{equation}
We used the Huber loss during training because we found that it behaves better for optimization when the predictions are quite close to the true values, and was otherwise an equivalent objective.


\section{Experimental setup}

Our soccer dataset consists of 2D ball/player positions from 95 professional soccer game halves mostly from the Bundesliga 2009-2010 competition, recorded at 10Hz (i.e. 0.1s between timesteps). It was collected using video-based player registration provided by AMISCO and is annotated with events like ``Pass'', ``Reception'' and ``Out for corner''. These events are used to partition the timesteps into ``in-play'' and ``out-of-play''. This data includes nominal roles for each player for each game half, which we used for canonical ordering.

We trained all models using PyTorch \cite{PyTorch}. Unless otherwise stated, we trained for 200 epochs with the training parameters described in \cite{Yeh2019}: the Adam optimizer \cite{Kingma2014} with an initial learning rate of 0.0005, and decaying by a factor of 0.999 after every epoch. We use every possible temporal window of player positions that are ``in-play'' with no more than 50\% overlap with the previous window. For data augmentation, we randomly flip examples horizontally and vertically.

Following \cite{Yeh2019, Felsen2018}, in all of our experiments, unless otherwise specified, we use a $T=50$ timesteps (5 second) temporal window of player positions, with $n=10$ timesteps (1 second) for the input and 40 timesteps (4 seconds) for the output.

There is an experiment in which we use 250 timesteps with $240$ timesteps (24 seconds) for the output; as this dramatically reduced the number of training examples, we trained for 400 epochs.

\section{Results} \label{sec:results}

We evaluate several baseline models and our GraN-MA architecture with sparse trajectories and (where trivially possible) full-trajectory conditioning. To compare with state-of-the-art, we train a RED \cite{Becker2018} model and a GVRNN \cite{Yeh2019} model on our dataset with and without sparse outputs. Neither RED nor GVRNN were adapted for full-trajectory conditioning because the changes required to include full trajectories and combine that information with the partial trajectories would be significant enough that they could no longer reasonably be called the same model. When training the RED model, we used a learning rate of 0.005 and no schedule, as specified in their paper. We also compare these to linear extrapolation (Lin. Ext.). We trained each model 7 times to account for performance variation due to random initialisation of weights, and report the mean and standard deviation of each setup. Unless otherwise stated, the shown number is the average L2 error over all examples, over all trajectories and over every time step therein. Unless otherwise stated, ``sparsity'' refers to how much time is assumed between predictions; this also indirectly describes the number of predictions actually made for our 4s window.

We implemented GVRNN based on the paper, and the authors of GVRNN provided model code which we incorporated into our training/testing code. We used the naive method of selecting sparse outputs from GVRNN. Unfortunately, neither implementation was able to train a model which produced meaningful results on our dataset, either with or without sparse outputs. We observed significant mode collapse \cite{Goodfellow2014} of a sort: although it modelled the distribution of acceleration and velocities of the players well, it always pushed all players towards one point on the field. Thus, on the L2 error metric, it always performed significantly worse than linear extrapolation.

\subsection{Varying Sparsity}

\begin{figure}
  \centering
  \includegraphics[width=0.45\textwidth]{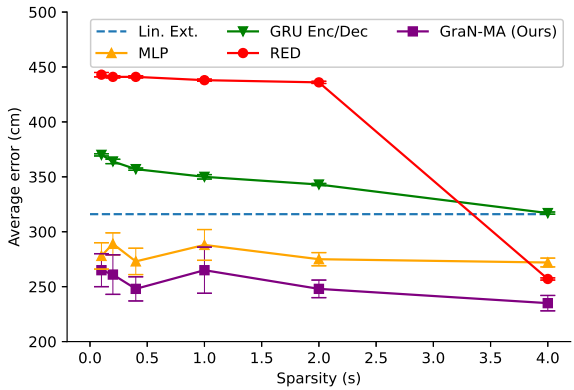}
  \caption{GraN-MA outperformed all baselines. And all models (except MLP) performed noticably better with a prediction sparsity of 4.0s. The RED model was only able to learn a good strategy when trained to predict one point (4.0s). The RED and the GRU encoder/decoder models had monotonic improvement with increasing sparsity. The models that performed better than Linear Extrapolation had much more variance in performance with more dense predictions. The error bar is the standard deviation across 7 random initialisations.}
  \label{fig:mainSparse}
\end{figure}

We trained several models with 0.1s (dense), 0.4s, 1.0s, 2.0s and 4.0s sparsity between each output. For the GRU encoder/decoder baseline and GraN-MA there was an obvious improvement when training with some sparse outputs, but there was no significant improvement for the MLP baseline (see Figure \ref{fig:mainSparse}). The GRU encoder/decoder model had monotonic improvement with increasing sparsity. RED had issues training and usually fell into predicting almost no motion. However, when trained with maximum sparsity (4.0s between outputs in this case), it learned to predict some motion and outperformed linear extrapolation and the baselines. We take this as evidence that the simplified output can help in cases where the model struggles to find anything but degenerate strategies. Recall that RED does not model human-human interactions. The GRU encoder/decoder and MLP baselines do model human-human interactions but were unable to use this extra information to outperform RED.

\subsection{Autoregressive models}

\begin{table}
\begin{center}
  \caption{The purely autoregressive models all did poorly on our dataset. As predicted, adding sparsity only made these autoregressive models worse. Mostly, they failed to converge to anything reasonable. All results are measured in cm, and the error is the standard deviation across 7 random initialisations.}
  \label{tab:autoregressive}
  \begin{tabular}{r c c c}
    \hline
    \textbf{Model} & $0.1s$ & $0.4s$ & $4.0s$ \\
    \hline
    Lin. Ext.      & $316$ {\tiny $\pm 00$} & - & -\\
    GraN-MA (Ours) & $265$ {\tiny $\pm 15$} & $248$ {\tiny $\pm 11$} & $\mathbf{235}$ {\tiny $\pm 07$} \\
    Simple GRU     & $400$ {\tiny $\pm 13$} & $3869$ {\tiny $\pm 1039$} & $550$ {\tiny $\pm 69$} \\
    Autoreg. CNN   & $539$ {\tiny $\pm 08$} & $2543$ {\tiny $\pm 1209$} & - \\
    GVRNN \cite{Yeh2019} & $1256$ & - & - \\
    \hline
  \end{tabular}
\end{center}
\end{table}

\begin{table*}
\begin{center}
  \caption{When trained/evaluated on a 24s prediction window, GraN-MA vastly outperforms linear extrapolation. Predicting a single point over 24s is worse than predicting every frame on such a large window. The best results here were obtained predicting once every 2.0s. All results are measured in cm, bold font is used to highlight the best result.}
  \label{tab:dtpVstp240}
  \begin{tabular}{r c c c c}
    \hline
    \textbf{Model} & $0.1s$ & $2.0s$ & $6.0s$ & $24.0s$ \\
    \hline
    Lin. Ext.      & $2683$ {\tiny $\pm 00$} & - & - & -\\
    GraN-MA (Ours) & $1129$ {\tiny $\pm 09$} & $\mathbf{1119}$ {\tiny $\pm 07$} & $1126$ {\tiny $\pm 08$} & $1279$ {\tiny $\pm 04$} \\
    \hline
  \end{tabular}
\end{center}
\end{table*}

\begin{table*}
\begin{center}
  \caption{Using full-trajectory conditioning dramatically improves performance, and this is further improved by sparse output predictions. All results are measured in cm, bold font is used to highlight the best results for each level of sparsity.}
  \label{tab:dtpVfxtp}
  \begin{tabular}{r c c c c}
    \hline
    &  & \multicolumn{3}{c}{\textbf{Full-trajectory conditioned}} \\
    \textbf{Model} & $0.1s$ & $0.1s$ & $1.0s$ & $4.0s$ \\
    \hline
    Lin. Ext.      & $278$ {\tiny $\pm 00$} & - & - & - \\
    Baseline MLP   & $254$ {\tiny $\pm 12$} & $229$ {\tiny $\pm 11$} & $218$ {\tiny $\pm 06$} & $202$ {\tiny $\pm 03$} \\
    GraN-MA (Ours) & $\mathbf{242}$ {\tiny $\pm 21$} & $\mathbf{158}$ {\tiny $\pm 06$} & $\mathbf{158}$ {\tiny $\pm 08$} & $\mathbf{149}$ {\tiny $\pm 01$} \\
    \hline
  \end{tabular}
\end{center}
\end{table*}

\begin{figure}
  \centering
  \includegraphics[width=0.45\textwidth]{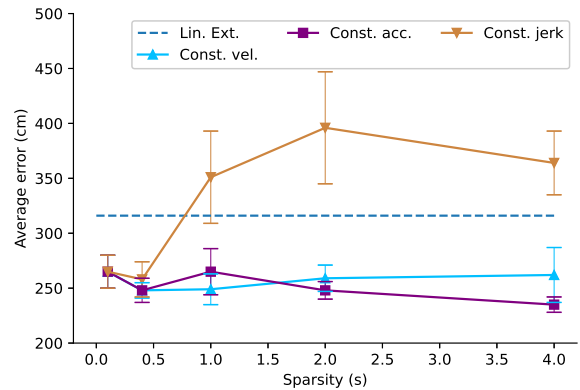}
  \caption{Constant acceleration is the best derivative to set to a constant to approximate the future positions for most training-time sparsities. Constant snap (N=4) performed so poorly that we did not show it here. The error bar is the standard deviation across 7 random initialisations.}
  \label{fig:constnthder}
\end{figure}

\begin{figure*}
  \centering
  \subfigure[Trained and evaluated on 4s window]{
    \includegraphics[width=0.45\textwidth]{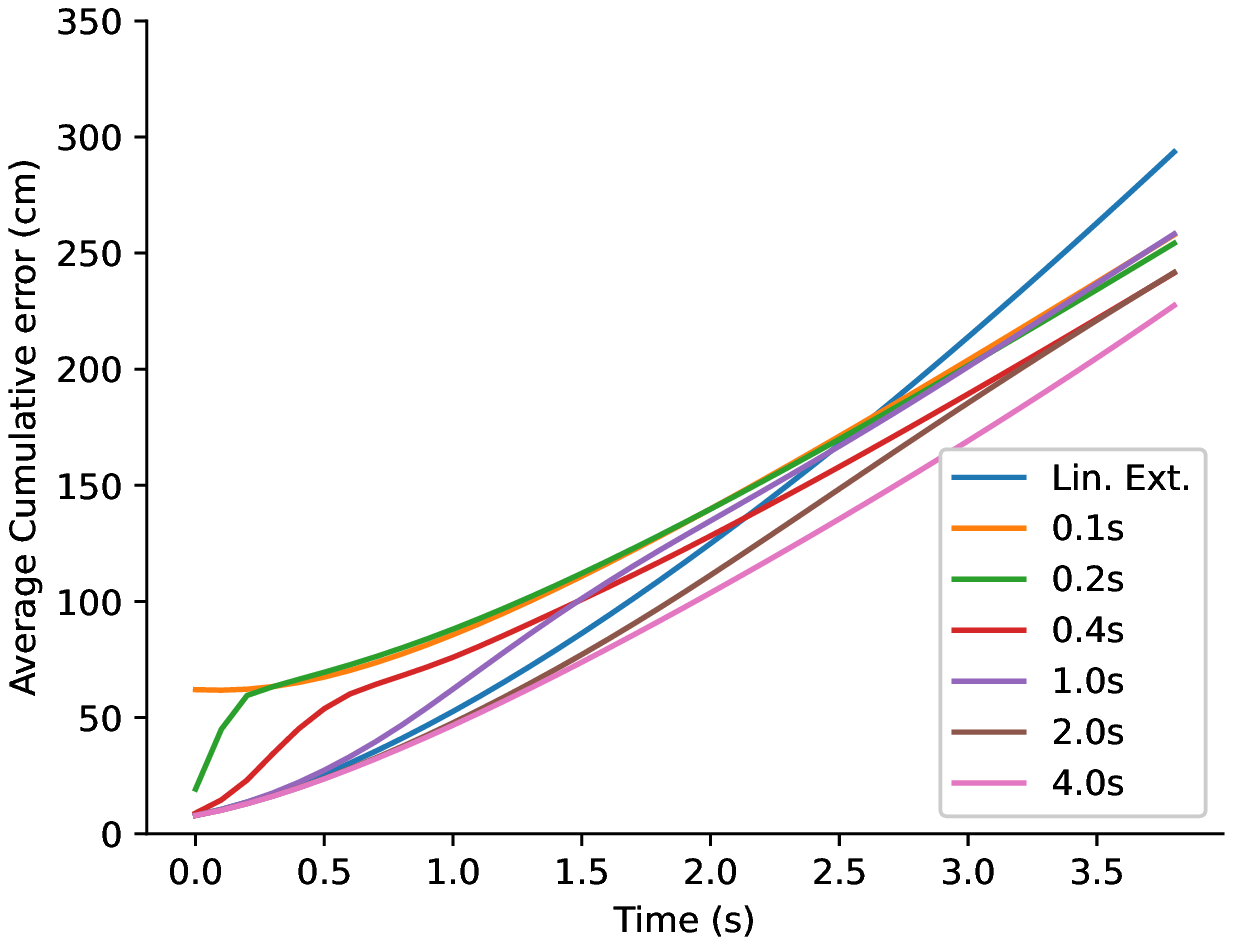}
  }
  \subfigure[Trained for 24s window, evaluated on 4s window]{
    \includegraphics[width=0.45\textwidth]{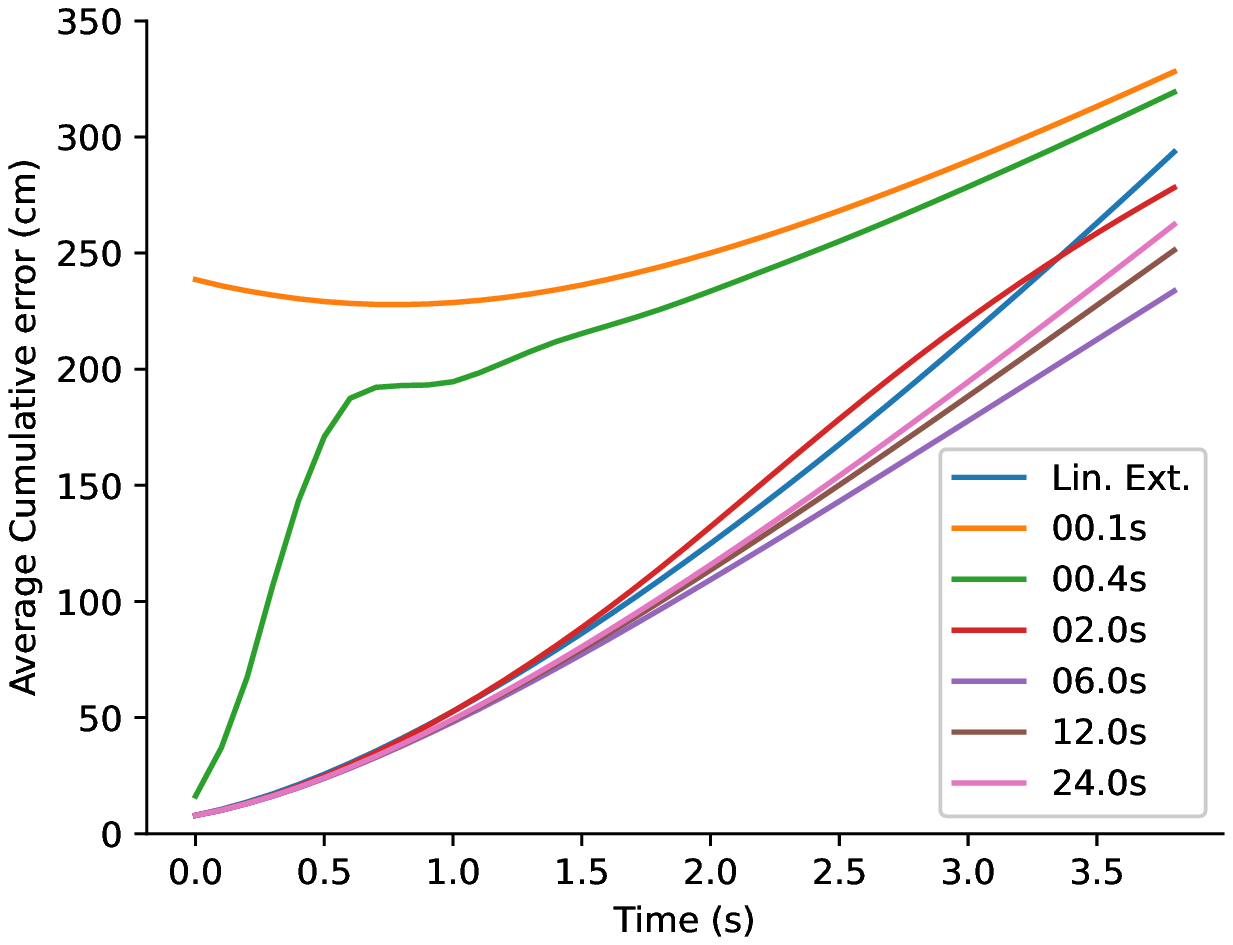}
  }
  \caption{We train GraN-MA at different sparsities (e.g. the 0.4s line is a GraN-MA model trained to produce outputs 0.4s apart) and on two prediction windows: 4s (\textbf{a}) and 24s (\textbf{b}). The average cumulative error is calculated as the average error on all time steps up to and including that time. Thus these graphs allow us to infer where, proportionally, the error is: early or late in the prediction window. e.g. The 0.1s (dense) model trained on a 4s window has much worse performance than linear extrapolation over the first second, but ends up with a better error when you include the whole 4s window; thus we know it does much better than linear extrapolation on later time steps to make up the difference. The early predictions for models trained with dense predictions on a 24s window are significantly worse than models trained for the 4s window.}
  \label{fig:trainWindowCompare}
\end{figure*}

We trained several autoregressive models with 0.1s (dense), 0.4s and 4.0s between each output (see Table \ref{tab:autoregressive}). These models were completely unable to converge to a useful strategy. The simple GRU and autoregressive CNN were trained with a much smaller learning rate ($10^{-6}$) to try to stabilise training (otherwise the losses jumped wildly), but they still did not converge to a reasonable solution. As predicted in Section \ref{sec:methodologyAutoRegressive}, using sparse outputs did not help these autoregressive models, and in fact actively hurt performance. We noted that for these autoregressive models sparse outputs would increase complexity of the output space. This has made it more difficult for these already poorly converging simple autoregressive models to converge. And the impact may not be as dramatic if these models had otherwise converged to a good solution.

\subsection{Constant N-th order derivative}

We trained GraN-MA models with constant velocity (N=1), acceleration (N=2), jerk (N=3) and snap (N=4) interpolation methods. We found that constant acceleration performed the best (see Figure \ref{fig:constnthder}). While constant velocity performed similarly to constant acceleration, constant jerk and snap performed significantly worse, with snap being as bad as the autoregressive models. Taking constant jerk as an example, rather than making a simple path between the points, there were often increasingly large deviations with larger sparsity and number of chained interpolations. When a human runs, the motion is smooth (near-constant velocity) for the majority of the time since any changes in direction occur in $<$0.5s. Enforcing a constant jerk also enforces a slowly-changing acceleration, which results in wide, unrealistic paths that oscillate around each prediction. We believe this to be the primary reason for the performance gap between constant acceleration and consant jerk (and higher order) interpolation approaches. We see reflected in increasingly large error with larger sparsity beyond 0.5s.

\subsection{Long time horizon}\label{sec:resultsLongTrain}

We found that models trained with dense outputs consistently performed worse than linear extrapolation on short time horizons ($<$2s into the future) when trained on both a 4s (40 outputs) and 24s (240 outputs) prediction window (see Figure \ref{fig:trainWindowCompare}). We found that for the larger prediction window size, GraN-MA's relative performance compared to linear extrapolation improved dramatically overall (see Table \ref{tab:dtpVstp240}). However, GraN-MA's relative performance was also much worse on short time horizons than when trained on a 4s prediction window. Then, the apparent improvement is just because linear extrapolation is so bad on longer time horizons. That is, it could sacrifice performance on early time steps and still massively outperform linear extrapolation by getting most of the later predictions not horrendously incorrect. These results suggests that, for soccer, a prediction sparsity between 2.0s and 4.0s is optimal.

\subsection{Full-trajectory conditioning}

We trained an MLP baseline and GraN-MA on full-trajectory conditioning. These models already outperformed linear extrapolation using dense prediction, and they performed even better on the full-trajectory conditioning task (see Table \ref{tab:dtpVfxtp}). GraN-MA performed dramatically better with the extra information as compared to the baseline MLP. This implies that GraN-MA is able to better utilise the extra information provided in the full trajectories. Additionally, the error was further improved by combining full-trajectory conditioning and sparse outputs. That is, sparse outputs is a complementary technique to training on a simpler task.

\subsection{Sparsity at evaluation time}\label{sec:resultsEvalSparsity}

\begin{figure}
  \hspace*{-2mm}
  \includegraphics[width=0.5\textwidth]{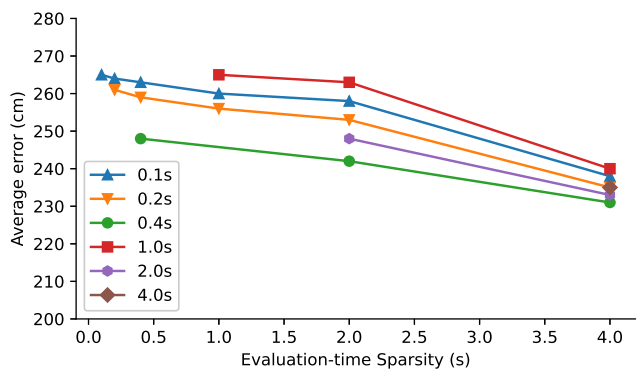}
  \caption{We train GraN-MA at different sparsities and then evaluate with increasing evaluation-time sparsity. We observe monotonic improvement with increasing evaluation-time sparsity.}
  \label{fig:sparseNLast}
\end{figure}

For the baseline MLP and GraN-MA, the error was not distributed over the output timesteps evenly (see Figure \ref{fig:trainWindowCompare}). In Figure \ref{fig:sparseNLast} we show the results of training separate GraN-MA models on different sparsities, and then evaluating by ignoring some of the outputs, effectively increasing the prediction sparsity at evaluation time only; e.g. given a model trained to produce outputs every 0.2s, selecting every 5th output evaluates that model as if it produced outputs only every 1.0s (interpolating between in all cases). For the range of values tested, there is a strictly monotonic improvement with increasing evaluation-time sparsity which we do not observe with increasing training sparsity. The strict monotonic improvement can be expected on any model that generally predicts better than linear extrapolation only at later time steps, since our constant acceleration interpolation is a smooth transition between linear extrapolation and the later prediction.

In all experiments there was an obvious and sharp improvement when evaluating at 4.0s of sparsity, which is a single prediction on our 4.0s evaluation window as compared to 2.0s of sparsity which is two predictions. Recall that constant acceleration can only change direction once for each prediction. This implies that the general direction that the player moves is --- with regards to L2 error --- both more predictable and more important than smaller movements.

\section{Conclusion}

Our results confirm the surprising phenomenon of deep learning models struggling to outperform linear extrapolation for trajectory prediction. To address this we have described a method utilising sparse outputs and constant acceleration interpolation to improve the training of said models. We have also described a novel architecture (GraN-MA) based on graph networks and multi-head attention for team-sport-based trajectory prediction and shown that it outperforms established state-of-the-art models. We show that constant acceleration interpolation can improve L2 error for GraN-MA and two simple baseline models. Further, we show that existing, non-degenerate deep learning models can be improved by only using a sparse subset of the predictions. Additionally, we show that for the application of soccer player trajectory prediction, a model with access to the full trajectories of a subset of the players (the attackers) performs significantly better, and that this is also further improved by training the model to predict sparse outputs.


Sparse trajectories are applicable to any trajectory prediction problem, so it would be useful to see if these improvements in performance are observed on other tasks and with other architectures, especially variational auto-encoders.

Although RNNs are considered the quintessential deep learning model for sequence modelling, we found that they typically performed worse than MLPs on our dataset. We believe it will be beneficial to explore the reasons for this discrepancy in future work and seek to improve RNNs specifically for predicting position data.

\begin{acknowledgements}
We would like to thank the Australin Institute of Sport (AIS) for funding this work.
And we would like to thank Stuart Morgan for organising the data and for consistently championing our Deep Learning research for sports at La Trobe.
\end{acknowledgements}

\section*{Conflict of interest}

The authors declare that they have no conflict of interest.

\bibliographystyle{abbrv}
\bibliography{Ghosting}

\end{document}